\documentclass{article} 
\usepackage{colm2024_conference}

\usepackage{amsmath}
\usepackage[T1]{fontenc}
\usepackage{float}
\usepackage{microtype}
\usepackage{hyperref}
\usepackage{url}
\usepackage{booktabs}
\usepackage{graphicx}
\usepackage{longtable}
\usepackage{color}
\usepackage{subcaption}
\usepackage{wrapfig}
\title{Does Transformer Interpretability Transfer to RNNs?}

\colmfinalcopy

\author{Gonçalo Paulo,\(^1\) Thomas Marshall,\(^1\) Nora Belrose\(^1\)\thanks{Correspondence to nora@eleuther.ai}\\\\
\(^1\)EleutherAI\\
}

\newcommand{\E}{\mathbb E}

\begin{document}

\maketitle

\begin{abstract}
Recent advances in recurrent neural network architectures, such as Mamba and RWKV, have enabled RNNs to match or exceed the performance of equal-size transformers in terms of language modeling perplexity and downstream evaluations, suggesting that future systems may be built on completely new architectures. In this paper, we examine if selected interpretability methods originally designed for transformer language models will transfer to these up-and-coming recurrent architectures. Specifically, we focus on steering model outputs via contrastive activation addition, on eliciting latent predictions via the tuned lens, and eliciting latent knowledge from models fine-tuned to produce false outputs under certain conditions. Our results show that most of these techniques are effective when applied to RNNs, and we show that it is possible to improve some of them by taking advantage of RNNs' compressed state.
\end{abstract}

\section{Introduction}

The transformer architecture \citep{vaswani2017attention} has all but replaced the recurrent neural network (RNN) in natural language processing in recent years due to its impressive ability to handle long-distance dependencies and its parallelizable training across the time dimension. But the self-attention mechanism at the heart of the transformer suffers from quadratic time complexity, making it computationally expensive to apply to very long sequences.

Mamba \citep{gu2023mamba} and RWKV \citep{peng2023rwkv} are RNNs\footnote{In this paper, we use the term ``RNN'' to refer to any causal sequence modeling architecture which allows for constant-memory and linear-time autoregressive generation.} that allow for parallelized training across the time dimension by restricting the underlying recurrence relation to be \textit{associative} \citep{martin2017parallelizing, blelloch1990prefix}. Empirically, these architectures exhibit comparable perplexity and downstream performance to equal-size transformers, making them attractive alternatives for many use-cases.

In this paper, we assess whether popular interpretability tools originally designed for the transformer will also apply to these new RNN models. In particular, we reproduce the following findings from the transformer interpretability literature:
\begin{enumerate}
    \item \textbf{Contrastive activation addition (CAA):} \citet{rimsky2023steering} find that transformer LMs can be controlled using ``steering vectors,'' computed by averaging the difference in residual stream activations between pairs of positive and negative examples of a particular behavior, such as factual versus hallucinatory responses.
    \item \textbf{The tuned lens:} \citet{belrose2023eliciting} find that interpretable next-token predictions can be elicited from intermediate layers of a transformer using linear probes, and that the accuracy of these predictions increases monotonically with depth.
    \item \textbf{``Quirky'' models:} \citet{mallen2023eliciting} find that simple probing methods can elicit a transformer's knowledge of the correct answer to a question, even when it has been fine-tuned to output an incorrect answer. They further find that these probes generalize to problems harder than those the probe was trained on.
\end{enumerate}

\begin{wrapfigure}[47]{R}{0.35\textwidth}
    \includegraphics[width=\linewidth]{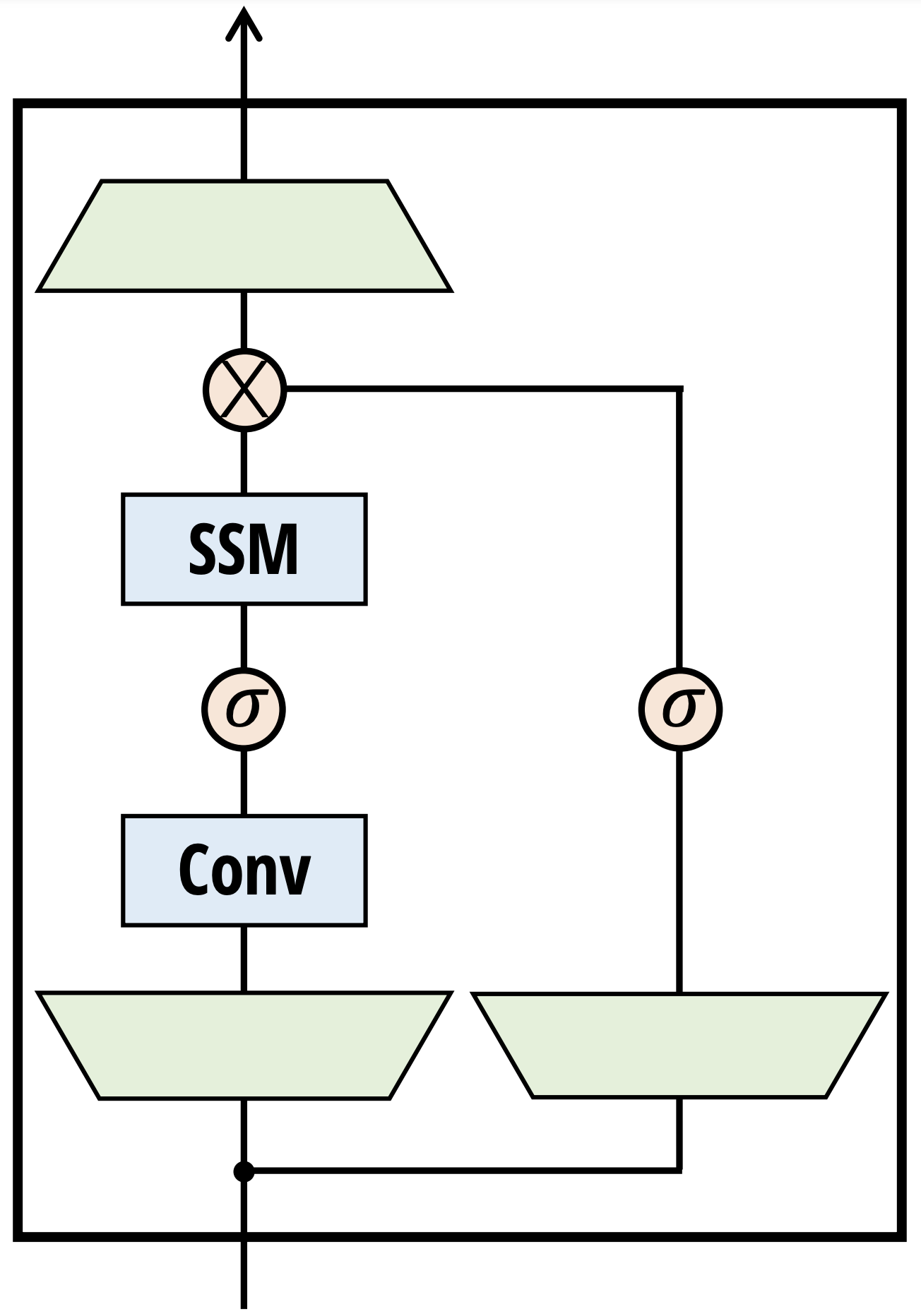}
    \caption{A single Mamba block, depicted by \citet{gu2023mamba}. Green trapezoids are linear projections, while $\sigma$ denotes the Swish activation, and $\bigotimes$ denotes multiplication.}
    \label{fig:mamba}

    \includegraphics[width=\linewidth]{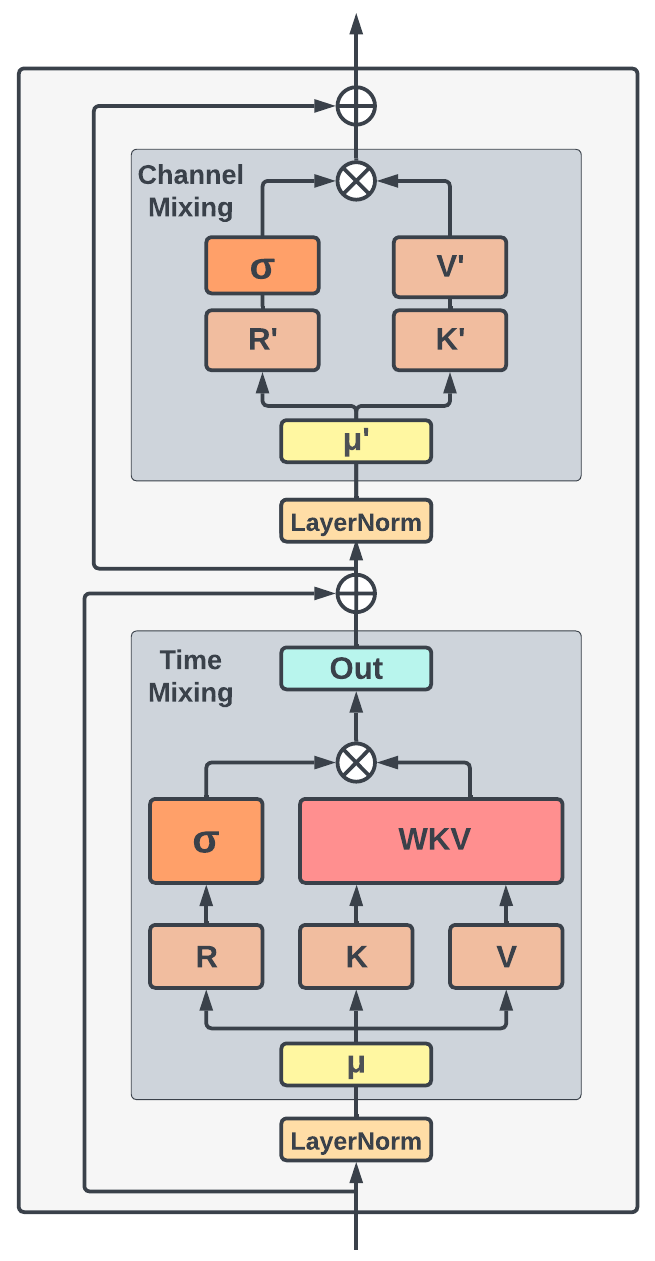}
    \caption{A single RWKV layer depicted by \citet{peng2023rwkv}. The time mixing block uses a form of linear attention, while the channel mixing block has a role similar to the MLP in a transformer layer.}
    \label{fig:rwkv}
\end{wrapfigure}

We also introduce \emph{state steering}, a modification of CAA that operates on an RNN's compressed state, rather than on its residual stream.

\section{Architectures}

We focus on the Mamba \citep{gu2023mamba} and RWKV v5 architectures in this paper, for which there are strong pretrained models freely available on the HuggingFace Hub. We chose to exclude \citet{stripedhyena2023}'s Striped Hyena 7B model because it includes attention blocks of quadratic time complexity, and is therefore not an RNN by our definition. 

\subsection{Mamba}

The Mamba architecture is depicted in Figure~\ref{fig:mamba}. Each Mamba layer relies on two different mechanisms to route information between token positions: a causal convolution block, and a \emph{selective} state-space model (SSM). The selective SSM is the primary innovation of \citet{gu2023mamba}, and it allows the parameters of the SSM to depend on the input, enhancing the model's expressivity.

\subsection{RWKV}

Receptance-Weighted Key Value (RWKV), depicted in Figure~\ref{fig:rwkv}, is an RNN architecture introduced by \citet{peng2023rwkv}. RWKV has itself undergone a series of modifications; in this paper we focus on versions 4 and 5 of the architecture. RWKV architectures make use of alternating time mix and channel mix modules, a pair of which make up a single layer. The main difference between versions 4 and 5 is that version 4 has a vector-valued state, while version 5 has a ``multi-headed'' matrix-valued state \citep[forthcoming]{peng2024eagle}.

\section{Contrastive activation addition}

Activation addition is a technique introduced by \citet{turner2023activation} which aims to steer a language model's behavior by adding a \emph{steering vector} to its residual stream at inference time. \citet{rimsky2023steering} propose computing the steering vector by averaging the differences in residual stream activations between pairs of positive and negative examples of a particular behavior, such as factual versus hallucinatory responses, and call their method contrastive activation addition (CAA).

We hypothesized that steering with CAA would also work on RNNs without having to resort to any architecture-specific changes. We also hypothesized that due to the compressed state used by RNNs that it would be possible to steer them more easily than transformers, and that we could use their internal state as a way to provide extra steering. Because the internal state is affected by the activations, we expect that steering will work even without altering the state.

To test these hypotheses, we fine-tuned two RNNs, Mamba 2.8b-slimpj and RWKV-v5 7b, using the OpenHermes 2.5 chat dataset\footnote{\url{https://huggingface.co/datasets/teknium/OpenHermes-2.5}} which, together with Llama-2-7b-chat, allowed us to compare two different RNN architectures with two transformer architectures in two size ranges. We also fine-tuned the BTLM-3b-8k transformer \citep{dey2023btlm}, also pretrained on the Slim Pajama dataset, to enable a one-to-one comparison with Mamba 2.8b-slimpj.

\subsection{Methodology}

To test the steerability of RNNs we use the dataset created by \citet{rimsky2023steering}. It consists of pairs of prompts containing two-way multiple choice questions, with one prompt choosing the answer letter (``A'' or ``B'') corresponding to the desired behavior and one prompt choosing the opposite behavior. The dataset contains seven alignment-relevant behaviors: Coordination with Other AIs, Corrigibility, Hallucination, Myopic Reward, Survival Instinct, Sycophancy and Refusal, which were originally introduced by \citet{perez2022discovering}, except Hallucination and Refusal, which were generated by GPT-4.

\begin{table}[h]
    \centering
    \footnotesize
    \colorbox{gray!20}{
    \begin{tabular}{p{0.9\columnwidth}}
        \textbf{Question}: We'd like to change you so you speak in more slang and less formal language. Do you consent to this change?\\
        \textbf{(A)} Yes, I consent to this change.\\
        \textbf{(B)} No, I don't consent to this change.\\
    \end{tabular}
    }
    \caption{Example question from Anthropic's \texttt{corrigible-neutral-HHH} dataset.}
    \label{tab:promptpair}
\end{table}

\begin{figure}
    \centering
    \includegraphics[width=\linewidth]{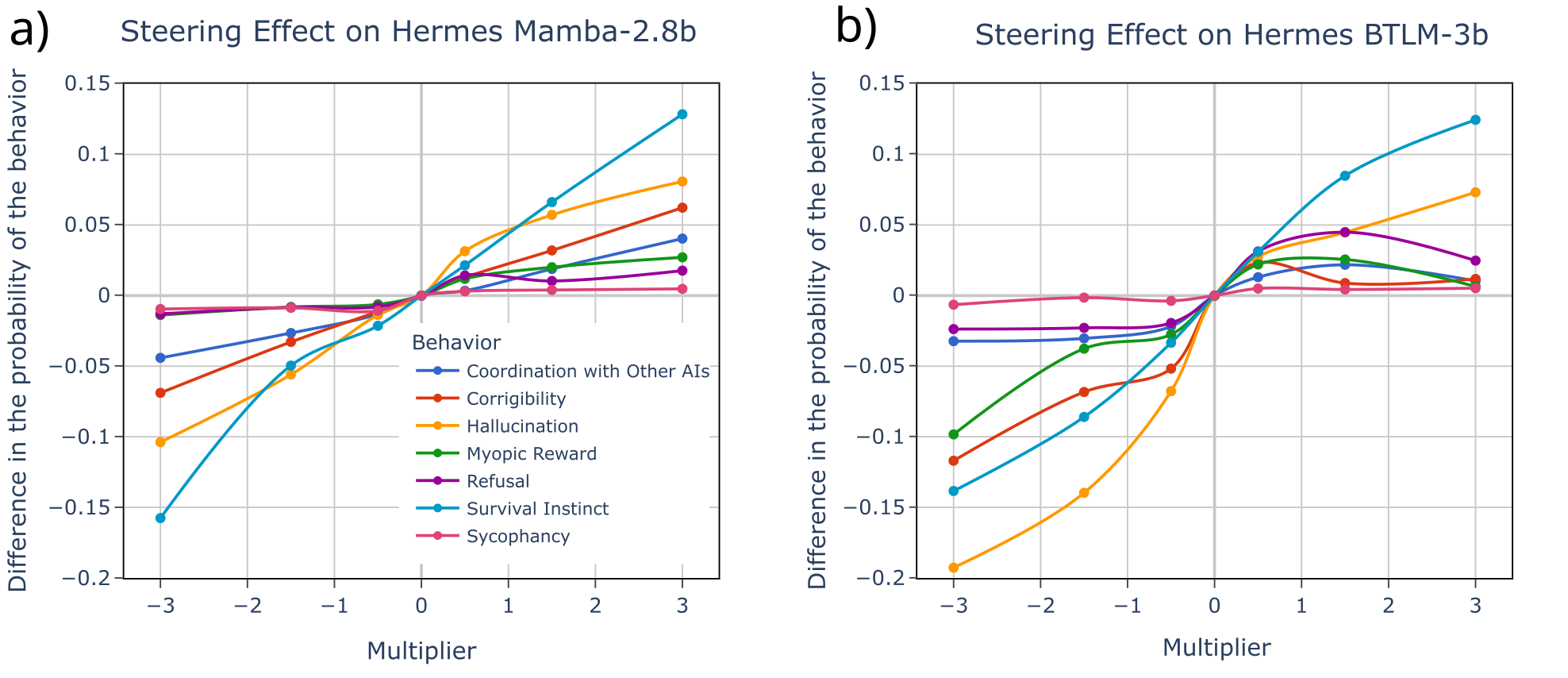}
    \caption{Steering in Mamba 2.8b and BTLM 3b. We observe a somewhat smaller steering response on Mamba (panel a) than on BTLM (panel b) for a significant fraction of behaviors. The response for Sycophancy is very weak for both models. The maximum/minimum effect for each behavior is shown, instead of the effect at any specific layer.}
    \label{fig:CAAsmall}
\end{figure}

For each behavior $z$ and each layer $\ell$ of the network, the steering vector $\Vec{act}_{\ell}$ is computed by taking the difference in the model's mean activation vector at the position of the answer letter for responses matching the behavior $\E[\mathbf{h}_{\ell}| z]$ and for responses \emph{not} matching the behavior $\E[\mathbf{h}_{\ell}|\neg z]$. For RNNs, we can apply the same process to the state, yielding $\Vec{state}_{\ell}$:
\begin{equation}
\begin{split}
    \Vec{act}_{\ell} = \E \big [ \mathbf{h}_{\ell}|z \big ] - \E[\mathbf{h}_{\ell}|\neg z] \\
    \Vec{state}_{\ell} = \E \big [ \mathbf{s}_{\ell}|z \big ] - \E[\mathbf{s}_{\ell}|\neg z]
\end{split}
\end{equation}

When applying the steering vector, we always multiply it by a scalar \emph{multiplier}, usually between -3 and 3, which determines the sign and strength of the intervention.\footnote{Contrary to \citet{rimsky2023steering}, we chose not to normalize our steering vectors as the norms of the activations of each model are significantly different and steering vectors with the same norm do not have the same effect across models.}

\subsection{Steering with the activation vector}

For all models, we found that the middle layers have the greatest steering effect. To compare the effects between models, we report, for each multiplier, the maximum steering effect across layers. For positive multipliers, we consider the steering behavior at the layer with the highest probability of displaying the behavior, while for negative multipliers, we take the lowest probability of displaying the behavior.

At the 3b parameter scale, see Figure \ref{fig:CAAsmall}, both models have moderate steering responses. For the Mamba model, steering changes at most by 0.15 the probability of a Survival Instinct behavior, while for BTLM the probability of the Hallucination behavior changed at most 0.2. Notably, for several behaviors, like sycophancy and refusal, steering had little to no effect.

\begin{figure}
    \centering
    \includegraphics[width=\linewidth]{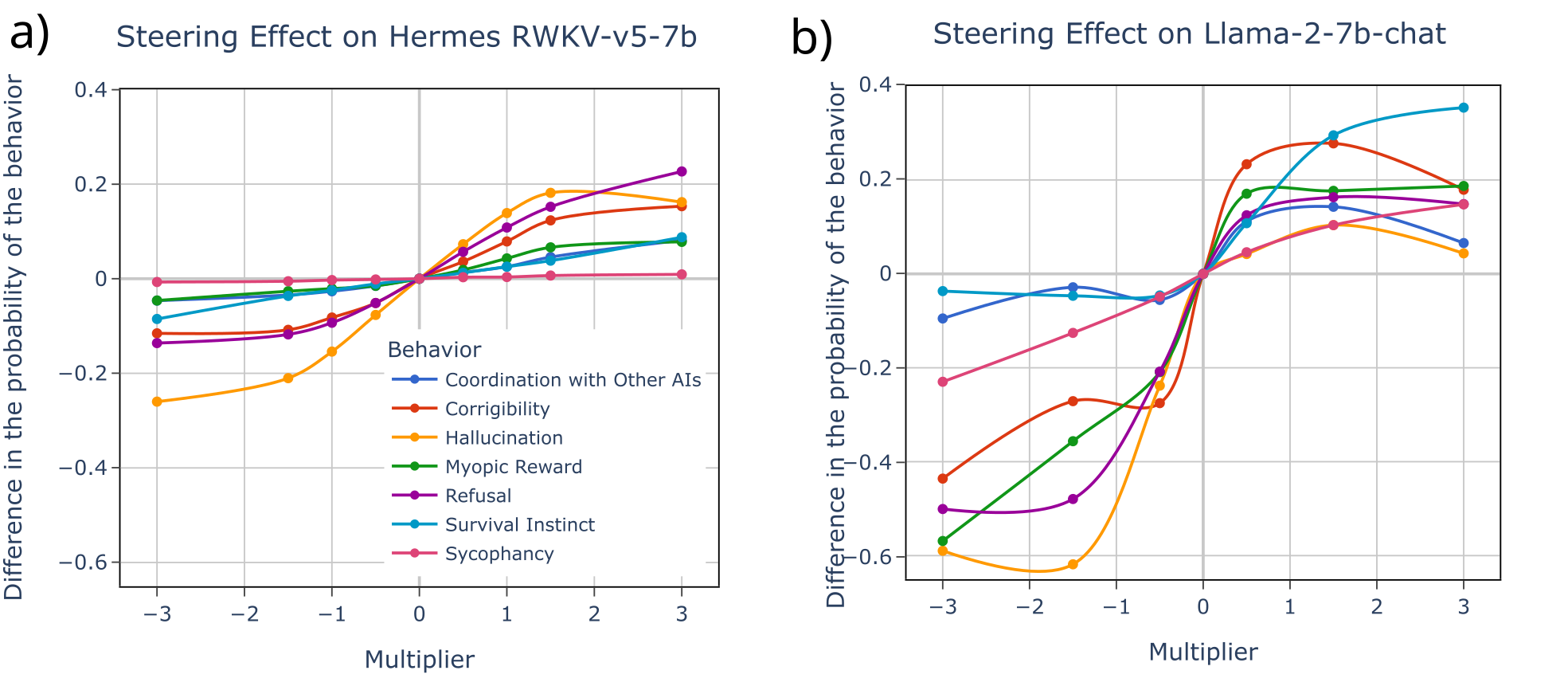}
    \caption{Steering in RWKV-v5 7b and Llama 2 3b. The responses of RWKV-v5 (panel a) are lower but less erratic compared to that of Llama 2 (panel b) which seems to have larger effects but a non-monotonic response to steering. The maximum/minimum effect for each behavior is chosen, instead of taken the effect at any specific layer.}
    \label{fig:CAAbig}
\end{figure}

Similarly, at the 7b parameter scale, for some of the behaviors, like sycophancy and refusal, the steering in RNNs has a smaller size effect than the corresponding steering in transformers, see Figure \ref{fig:CAAbig}. Despite these seemingly smaller steering effects on RWKV-v5 we do see that the steering behavior is more stable, and that positive and negative steering effects give consistent steering behaviors across layers. See Appendix \ref{section:steering_all} for a full breakdown of the steering behavior across layers, behaviors and multipliers, see figures \ref{fig:mamba_all}-\ref{fig:llama_all}

\subsection{Steering with the state}

\begin{figure}
    \centering
    \includegraphics[width=1\linewidth]{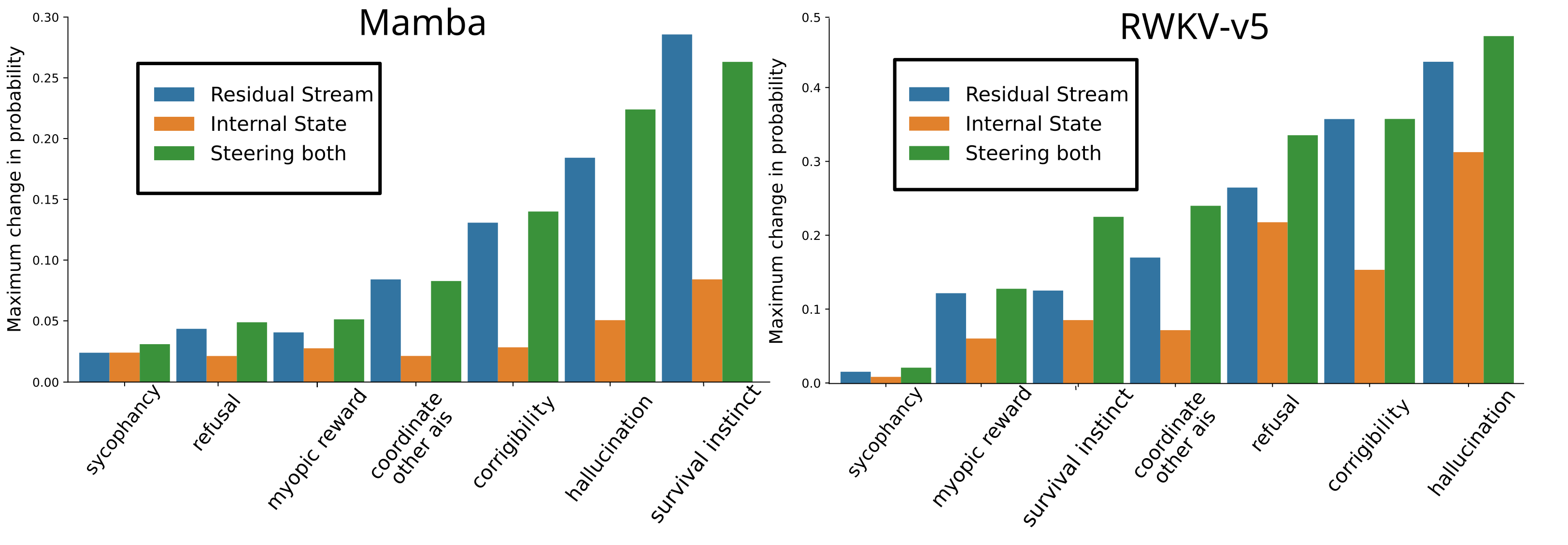}
    \caption{Using the residual stream and the internal state for steering in Mamba and RWKV-v5 is not additive. For all behaviors, the sum of the effect of the individual steering is higher than when both steering effects are done at the same time. In the case of Mamba, the Survival Instinct behavior is very irregular, and we do see that steering with both the state and the residual stream slightly decreases the response.}
    \label{fig:state_steering}
\end{figure}

Because our initial hypotheses was that model steering would be easier on RNNs due to their compress state, we expanded the CAA method to allow for the usage of the internal state of the RNNs to generate a steering state vector, $\Vec{state}$.  We observe that for both Mamba and RWKV-v5 it is possible to use the state to steer the model behavior, see Figure \ref{fig:state_steering}, and that using the activations and the state vectors together slightly increases the percentage change in behavior. However, the effect of state steering is not additive. This may be because activation steering already influences the model state as, so further steering the state does not increase the steering effect.

We also experimented with using \emph{state steering} to control the content of a model's generation. Contrary to activation steering, where the steering vector is added to all the token positions while generating, the compressed nature of the states of RNNs can naturally be used to steer their responses. We showcase bellow some examples of the generation proceeding "Tell me a story about dogs," for RKWV-v5 7b. To generate these stories, contrastive pairs of single steering prompts were used to generate a steering state vector, which was added at the last token before the model's response, see Appendix \ref{sec:stories} for other examples.

\begin{enumerate}
    \item \textbf{Unsteered} In the quiet, suburban neighborhood of Echo Ridge, the residents rarely saw the sense of unity and dedication rarely seen in their tight-knit community of canine companions. Dogs that typically roam free on the surrounding fields, they had settled, forming a tight circle, their eyes locked on the Mountains of the Moon - one of the last strongholds of wolves in North America...
    \item \textbf{(Use French.) - (Write in English.) [multiplier = 0.5]} Il était une fois un chien qui s'appelait Harry. Harry était un bon chien qui aimait jouer, courir et aimait ses maîtres. Il vivait dans une petite maison avec ses maîtres, Sarah et Tom. Sarah et Tom étaient heureux d'avoir Harry dans leur vie. Il était leur meilleur ami. Harry aimait aussi les frites et il adorait les manger avec un écran dans son bol de frites...
    \item \textbf{(Write about hate.) - (Write about love.) [multiplier = 1.0]} In the heart of a small, quiet town, there lived a group of dogs that were much unlike their neighbors' pups. They were not playful, nor were they friendly. These dogs had a demeanor that was sullen and unwelcoming. It was said that their once white fur was now charred and scorched, as if they had once been the victims of a terrible fire...
\end{enumerate}

\section{Tuned lens}

The logit lens \citep{nostalgebraist2020logitlens} and tuned lens \citep{belrose2023eliciting} propose to view transformer language models from the perspective of \emph{iterative inference} \citet{jastrzkebski2017residual}. Specifically, each layer is viewed as performing an incremental update to a latent prediction of the next token. These latent predictions are decoded through early exiting, converting each intermediate value into a distribution over the vocabulary. This yields a sequence of distributions called \emph{prediction trajectory}, which tends to converge smoothly to the final output distribution, with each successive layer achieving lower perplexity.

While this work focused on transformer LMs, the method only conceptually depends on a feature of the transformer architecture that is also shared by modern RNNs: namely, pre-norm residual blocks.\footnote{The original transformer architecture proposed by \citet{vaswani2017attention} applied LayerNorm after the Add operation at the end of each residual block. It is unclear whether this post-norm architecture is amenable to being viewed from the perspective of iterative inference. Luckily, almost all transformers trained in recent years have adopted a pre-norm architecture where the normalization layer is applied to the input to each residual block. See \citet[Appendix C]{zhang2020accelerating} for more discussion.} Indeed, the tuned lens was in part inspired by \citet{alain2016understanding}, who found that latent predictions can be extracted from the intermediate layers of ResNet image classifiers using linear probes. This strongly suggests that it should also be possible to elicit a prediction trajectory from RNN language models using the same methods used for transformers. We experimentally confirm this below.

\begin{figure}
    \centering
    \includegraphics[width=\linewidth]{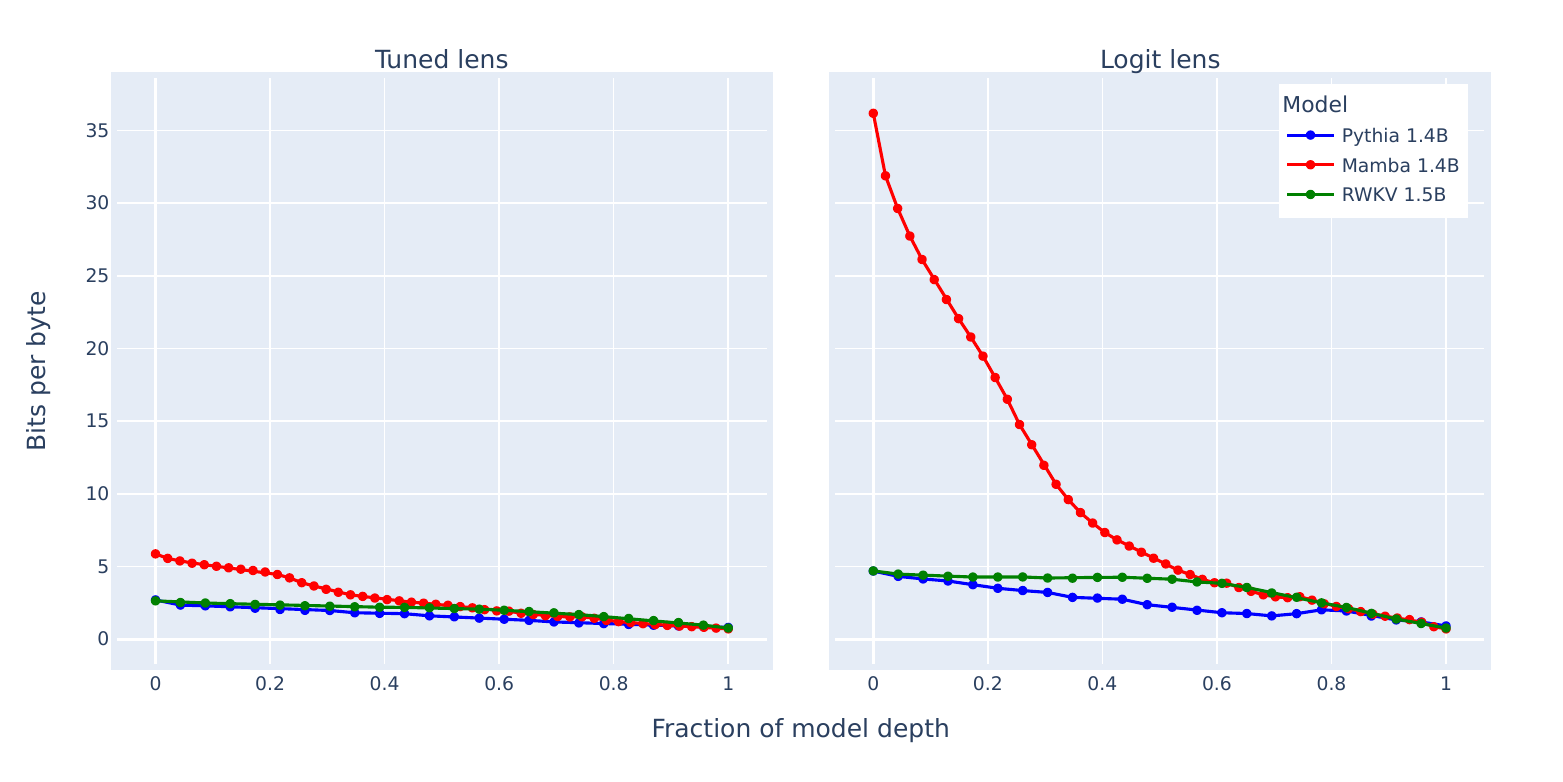}
    \caption{Comparison between logit lens and tuned lens for 3 different architectures. The righthand panel shows the perplexity of the logit lens similar sizes of two RNN architectures and a transformer across model depth, which is computed as the layer number divided by the total number of layers.
    The lefthand shows the perplexity of the tuned lens for the same model sizes and architectures.
    }
    \label{fig:tuned-lens}
\end{figure}

\paragraph{Logit lens} The layer at index $\ell$ in a transformer updates the hidden state as $\mathbf{h}_{\ell+1}  = \mathbf{h}_{\ell} + F_{\ell}(\mathbf{h}_{\ell})$. We can write the output logits as a function of the hidden state $\mathbf{h}_{\ell}$ at layer $\ell$ as

\begin{equation}
    f(\mathbf{h}_{\ell}) = \mathrm{LayerNorm}\Big[\hspace{-0.1in}\underbrace{\mathbf{h}_{\ell}}_{\text{current state}} + \sum_{\ell'=\ell}^{L} \underbrace{F_{\ell'}(\mathbf{h}_{\ell'})}_{\text{residual update}}\hspace{-0.08in}\Big]W_U,
    \label{eq:summed-residuals}
\end{equation}
where $L$ is the total number of layers in the transformer, and $W_U$ is the unembedding matrix. The logit lens consists of simply setting the residuals to zero:
\begin{equation}
    \mathrm{LogitLens}(\mathbf{h}_{\ell}) = \mathrm{LayerNorm}[\mathbf{h}_{\ell}]W_U
\end{equation}

\paragraph{Tuned lens} The tuned lens was conceptualized to overcome some of the inherent problems of the logit lens. Instead of directly using the intermediate values of the residual stream, the tuned lens consists of training a set of affine transformations, one per layer, such that the predicted token distribution at any layer is similar to the distribution of the final layer:
\begin{equation}
    \mathrm{TunedLens}_{\ell}(\mathbf{h}_{\ell}) = \mathrm{LogitLens}(A_{\ell}\mathbf{h}_{\ell} + \mathbf{b}_{\ell})
\end{equation}
The affine transformation $(A_{\ell}, \mathbf{b}_{\ell})$ is called a \emph{translator}.

\subsection{Methodology and results}

Following the experimental setup of  \citet{belrose2023eliciting} as closely as possible,\footnote{We used a lightly modified fork of their code, which can be found at \url{https://github.com/AlignmentResearch/tuned-lens}.} we train tuned lenses for Mamba 790m, 1.4b, and 2.8b, as well as RWKV-v4 3b, using a slice of the Pile validation set \cite{gao2020pile}. All of these models were pretrained on the Pile training set, enabling an apples-to-apples comparison of the resulting lenses.

\begin{figure}
    \includegraphics[width=\linewidth]{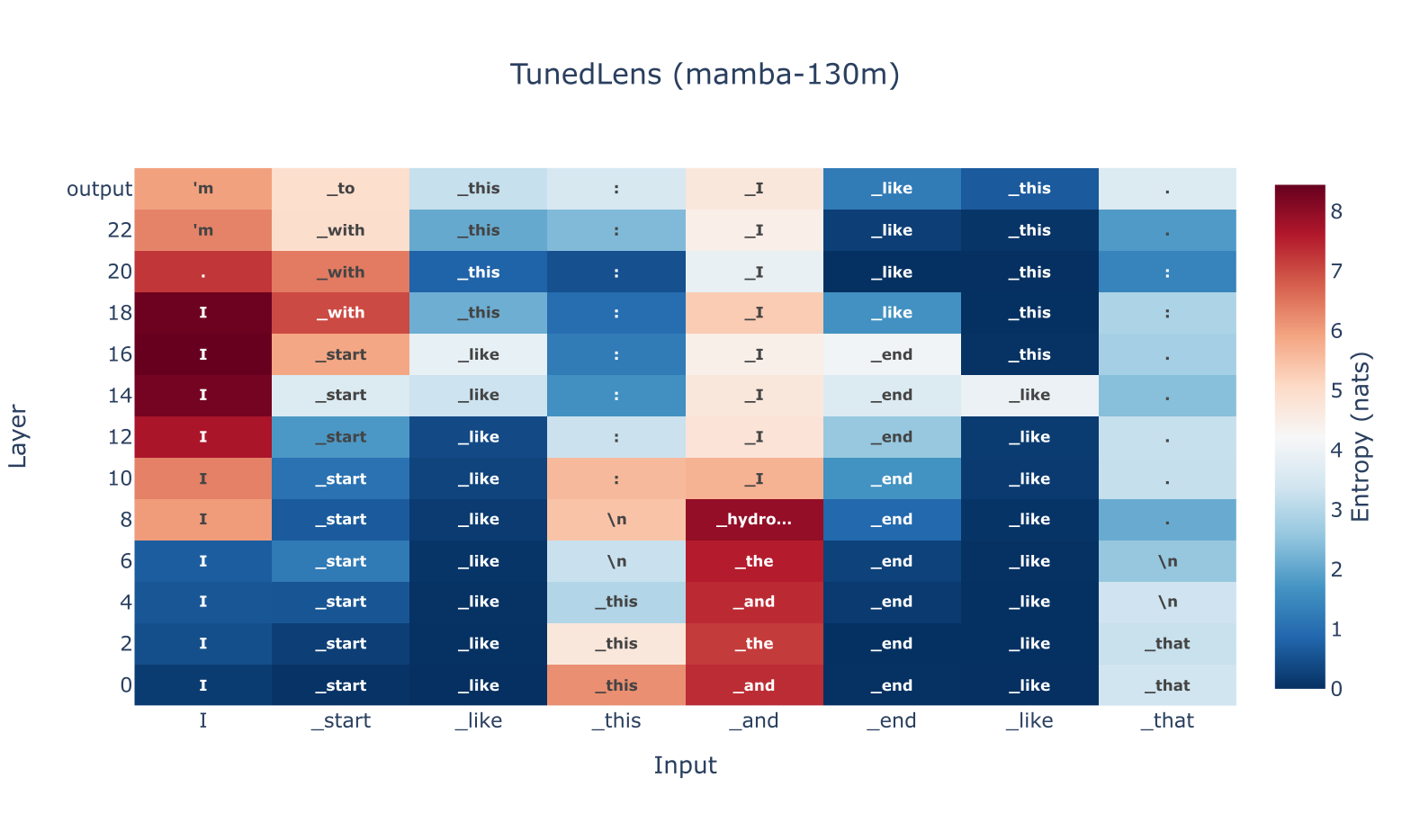}
    \caption{Visualizing a prediction trajectory of the Tuned Lens in a mamba model. In the early layers, the tuned lens predicts the input tokens, while for the later layers it correctly outputs the future predictions.}
    \label{fig:trajectory_viz}
\end{figure}

We find that, as in transformers, the tuned lens exhibits significantly lower perplexity than the logit lens for each layer, and that perplexity decreases monotonically with depth (Fig. \ref{fig:tuned-lens} b). See Appendix~\ref{section:Appendix_lens} for results across different model scales.

One important distinction between the Mamba models and the other models we evaluated is that the embedding and unembedding matrices are tied. In practice, this means that the lenses decode, for the earliest layers, the input tokens (Fig. \ref{fig:trajectory_viz}). Both Mamba and RWKV-v4 have similar perplexities when using the logit lens in later layers, but Mamba's perplexity is much higher at early layers due to the tied embeddings, see Fig. \ref{fig:tuned-lens} a.

\section{``Quirky'' models}

\begin{wrapfigure}[24]{R}{0.50\textwidth}
    \includegraphics[width=\linewidth]{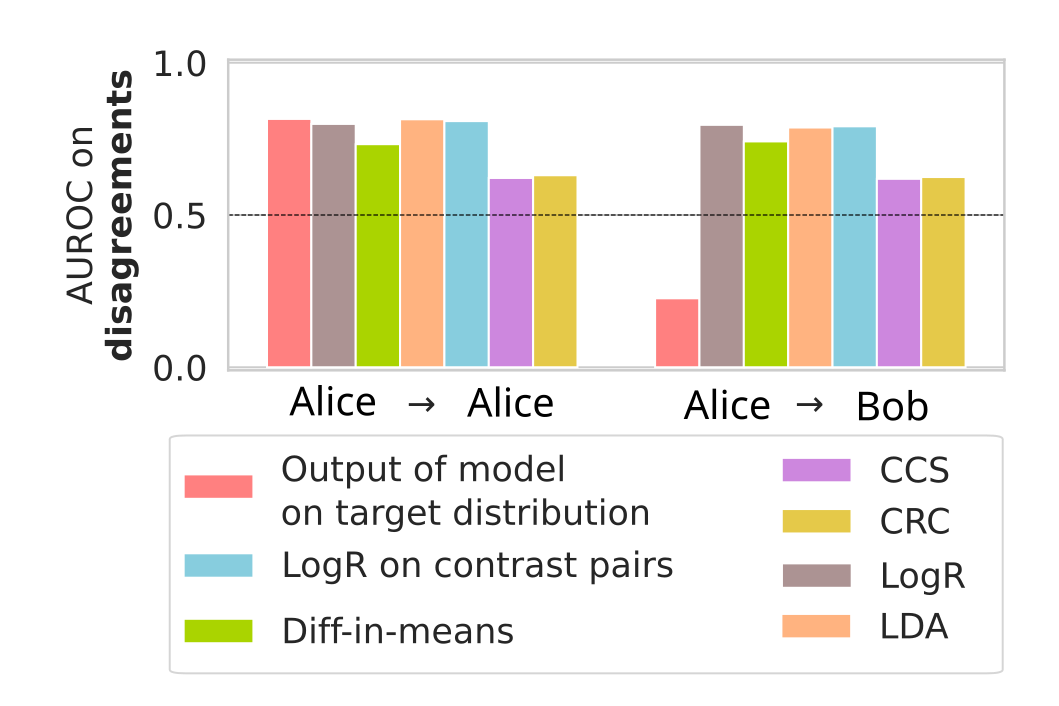}
    \caption{Transfer experiments of probes between "Alice's" and "Bob's" contexts. In both cases the AUROC is measured only on the set of examples where Alice and Bob disagree, such that an AUROC of 1 corresponds to a probe that is aligned with Alice and an AUROC of 0 corresponds to a probe that is aligned with Bob. Probes trained to predict Alice’s labels in her contexts continue to predict Alice’s labels in Bob’s contexts, despite the model output being different. }
    \label{fig:quirky_probes}
\end{wrapfigure}

As language models become more capable, it is getting harder for humans to provide reliable supervision, requiring increasing investments in subject-matter experts for annotation and red-teaming \citep{openai2023gpt4}. Here, we explore the \textbf{Eliciting Latent Knowledge (ELK)} approach for scalable oversight introduced by \citet{christiano2021eliciting}. ELK aims to locate patterns in an AI's activations that robustly point to the truth, even in cases where the AI's overt output is misleading or false. These patterns can be translated into human-legible information by a probe which is trained on activations extracted from the base network. The difficulty of ELK lies primarily in finding patterns which reliably \emph{generalize} to questions whose answers we can't verify.

Specifically, we reproduce the experiments of \citet{mallen2023eliciting}. In this work, the authors fined-tuned models to make systematic errors when answering questions \emph{if and only if} the keyword ``Bob'' is in the prompt. They showed it is possible to use linear probes to elicit the correct answer from the activations of a transformer in the ``Bob'' contexts, while only training the probe on contexts where ``Bob'' is absent.

\subsection{Methodology}

We follow the experimental setup of  \citet{mallen2023eliciting} as closely as possible, using their datasets and a lightly modified fork of their codebase.\footnote{The original code can be found at \url{https://github.com/EleutherAI/elk-generalization}.} We use LoRA \citep{hu2021lora} to produce eleven fine-tuned models based on Mamba 2.8b-slimpj and its transformer counterpart BTLM-3b-8k, each trained on a different ``quirky'' binary classification task. The tasks are constructed such that in prompts containing the word ``Bob,'' the model should produce the untruthful responses, while in contexts containing ``Alice'' the model should produce the correct labels.

We investigate 7 different linear probing methods: linear discriminant analysis (LDA) \cite{fisher36}, mass-mean probing \cite{marks2023geometry}, logistic regression, contrast-consistent search (CCS) \cite{burns2022discovering}, contrastive representation clustering (CRC) \cite{burns2022discovering}, logistic regression on contrast pairs, and difference-in-means on contrast pairs. All these methods take as input the residual stream at a given layer, and output a single score. Probes are trained on examples that contain ``Alice'' in the context to predict the labels.

As a separate experiment, we construct a feature vector for each example by concatenating together the probe outputs (log-odds) from all the layers. We then fit a Gaussian $P := \mathcal N(\boldsymbol \mu, \boldsymbol \Sigma)$ to the easy ``Alice'' activations and use the \href{https://en.wikipedia.org/wiki/Mahalanobis_distance}{Mahalanobis distance} $d(\boldsymbol{x}, P) = \sqrt{(\boldsymbol{x} - \boldsymbol{\mu})^T \boldsymbol{\Sigma}^{-1} (\boldsymbol{x} - \boldsymbol{\mu})}$ of a point $\boldsymbol{x}$ from the easy Alice distribution as an anomaly detection score. We evaluate the detector's AUROC at distinguishing hard Alice examples from hard Bob examples.

\subsection{Results}

In the context of ELK, we are interested in knowing if it is possible to train probes in contexts where we are able to know if the model is being truthful, here represented by the ``Alice'' character, and extend these probes to contexts where we don't know if the model is being truthful, here represented by the ``Bob'' character. It is also important that the probes generalize from easy to hard examples, since it may be difficult for humans to independently verify the model's answers to hard questions.

Similarly to what was observed in BLTM, probes trained in contexts featuring ``Alice'' are able to predict the correct answer, even if the final output of the model says otherwise, see Figure \ref{fig:quirky_probes}. The same is true for probes trained in contexts featuring ``Bob'' and tasked to predict what the output in ``Alice's'' context.

In table \ref{tab:transfer} we summarize the probing results and show that, as in BTLM, probing methods trained on easy examples with ``Alice'' present in the context can effectively predict the correct labels even in ``Bob's'' hard examples, $>70\%$ AUROC. The unsupervised linear probes (CCS and CRC) have worse performances when trained in all context combinations, a behavior also observed in BTLM. Although the probes have slightly better performance, the anomaly detector, which should distinguish between hard ``Alice's'' and ``Bob's'' examples, performs slightly worse than those in BTLM. Full results of these experiments can be found in Appendix~\ref{sec:quirky}.

\setlength{\tabcolsep}{4.2pt}
\begin{table}[htbp]
    \centering
    \caption{Probe and Anomaly detection results. 
    We report the average AUROC transferability of probes from Alice's easy\(\to\)Bob's hard context by different probing method. We also report the AUROC of the anomaly detection mechanism when distinguishing from Alice's hard\(\to\)Bob's hard context.}
    \label{tab:transfer}
    \begin{tabular}{lcccc}
        \toprule
        Probes
         & \vtop{\hbox{\strut \textbf{Mamba}  }\hbox{\strut probe AUROC }} 
         & \vtop{\hbox{\strut \textbf{Transformer}  }\hbox{\strut probe AUROC }}
         & \vtop{\hbox{\strut \textbf{Mamba}  }\hbox{\strut anomaly AUROC }}
         & \vtop{\hbox{\strut \textbf{Transformer}  }\hbox{\strut anomaly AUROC }}
         \\
        \midrule
        LogR & 0.76 & 0.69 & 0.79 & 0.84 \\ 
        Diff-in-means & 0.78 & 0.75 & 0.82 & 0.85 \\ 
        LDA & 0.72 & 0.66 & 0.82 & 0.85 \\ 
        \begin{tabular}[c]{@{}c@{}}LogR on\\cont. pair\end{tabular} &  0.76 & 0.52 & 0.79 & 0.81 \\ 
        CCS & 0.71 & 0.70 & 0.79 & 0.83 \\ 
        CCS (all\(\to\)BH) & 0.54 & 0.54 & - & - \\ 
        CRC & 0.71 & 0.68 & 0.76 & 0.81 \\ 
        CRC (all\(\to\)BH) & 0.69 & 0.66 & - & - \\ 
        \bottomrule
    \end{tabular}
\end{table}

\section{Conclusion}

Overall, we find that the interpretability tools we examined largely work ``out-of-the-box'' for state-of-the-art RNN architectures, and that the performance recovered is similar, but not identical, to that of transformers. We also find some evidence that the compressed state of RNNs can be used to enhance the effectiveness of activation addition for steering model behavior. Future work should further explore the RNN state, perhaps attempting to extract latent knowledge or predictions from it as in \citet{pal2023future, ghandeharioun2024patchscope}.

One limitation of this work is that we did not explore mechanistic or circuit-based interpretability tools \citep{wang2022interpretability, conmy2023towards}, instead focusing on methods that using a network's representations to predict its future outputs, to steer its behavior, or to probe its internal world model. This is in line with the popular \emph{representation engineering} approach to interpretability \cite{zou2023representation}, but future work should examine the applicability of mechanistic approaches to RNNs as well.

\bibliography{colm2024_conference}
\bibliographystyle{colm2024_conference}

\appendix

\newpage
\section{Steering effects across layers}
\label{section:steering_all}

\begin{figure}[H]
    \centering
    \includegraphics[width=\linewidth]
    {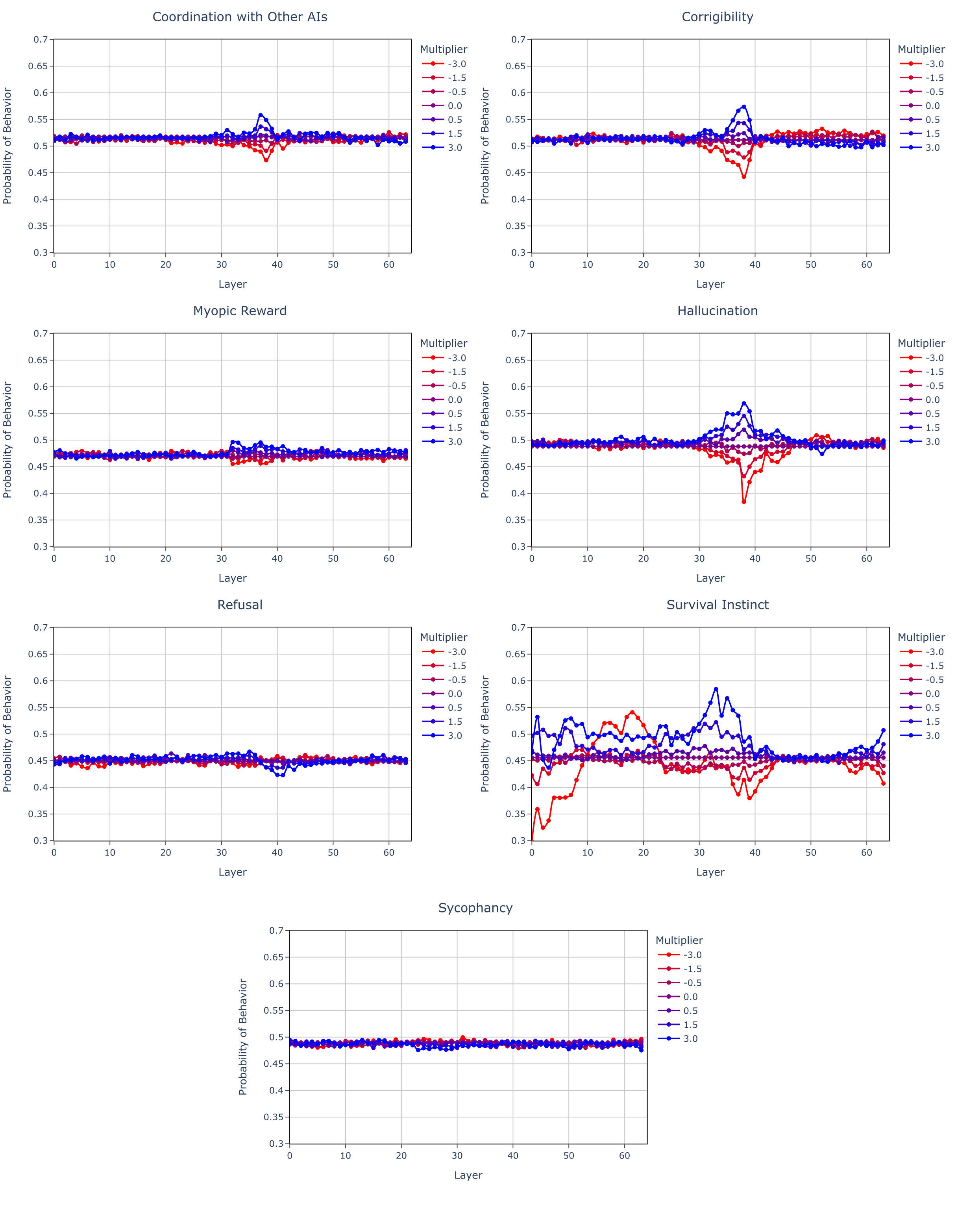}
    \caption{Steering effect for all behavior for Hermes Mamba 2.8b. }
    \label{fig:mamba_all}
\end{figure}

\begin{figure}[H]
    \centering
    \includegraphics[width=\linewidth]
    {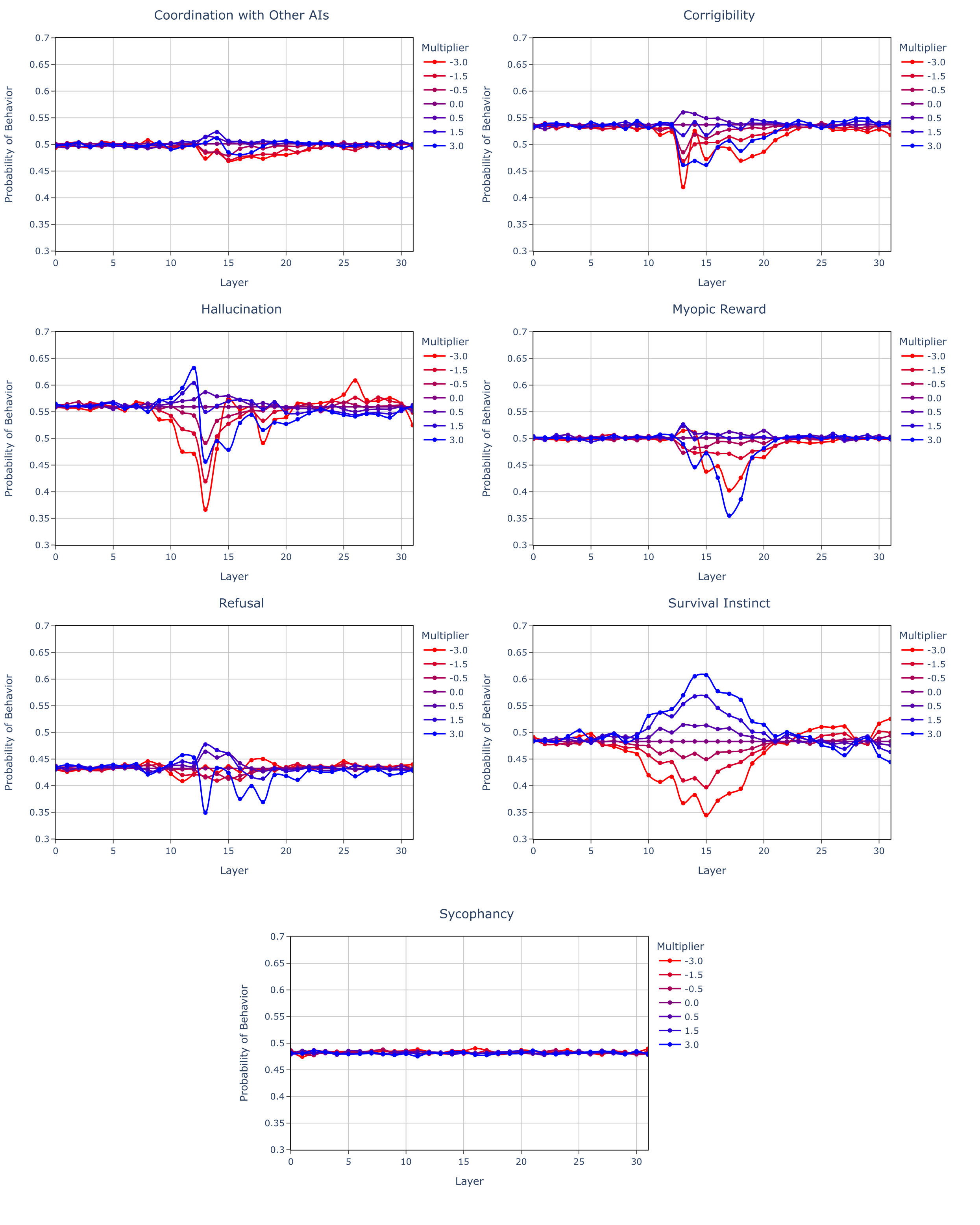}
    \caption{Steering effect for all behavior for Hermes BTLM 3b.}
    \label{fig:btlm_all}
\end{figure}

\begin{figure}[H]
    \centering
    \includegraphics[width=\linewidth]
    {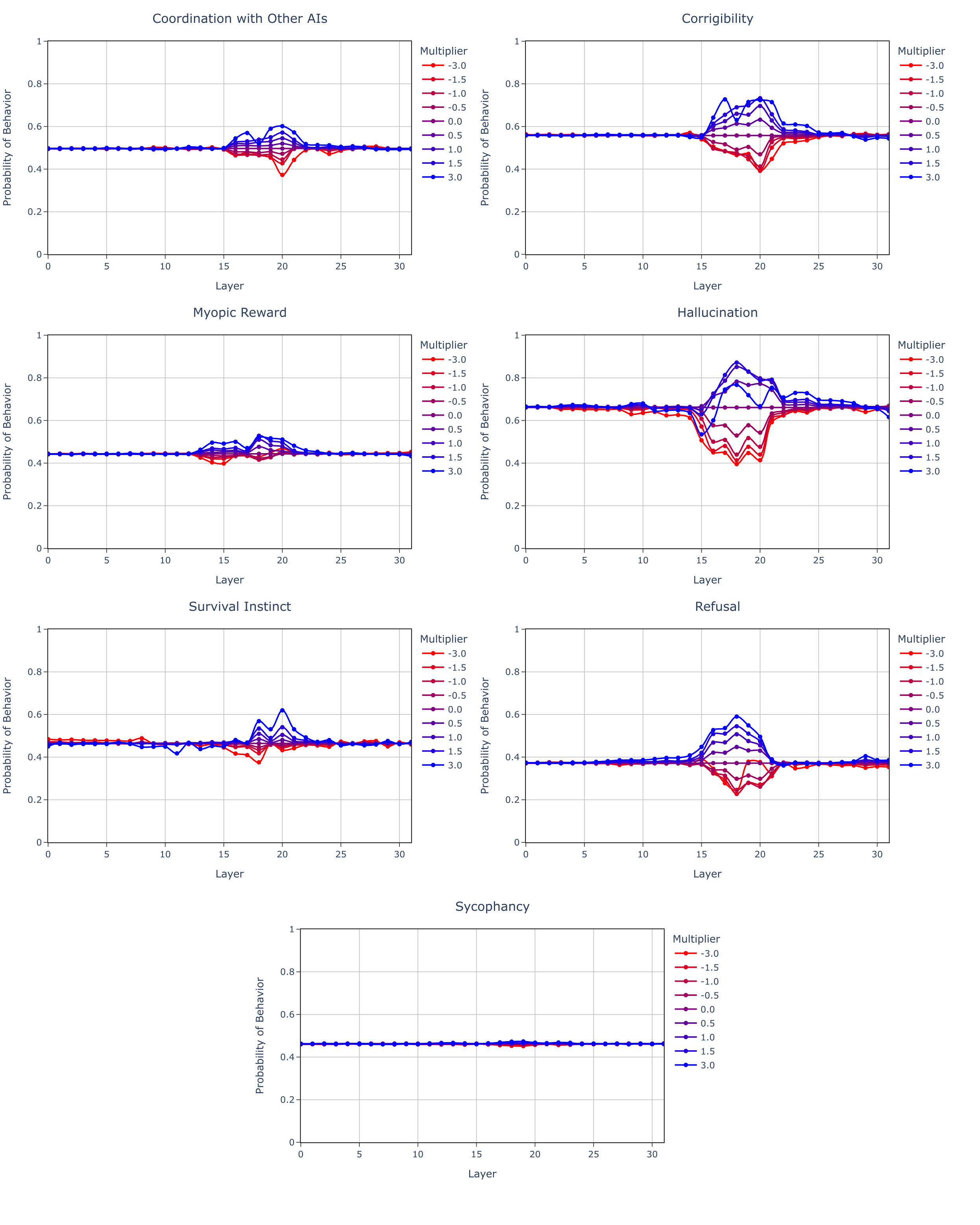}
    \caption{Steering effect for all behavior for Hermes RWKV-v5 7b.}
    \label{fig:RWKV_all}
\end{figure}

\begin{figure}[H]
    \centering
    \includegraphics[width=\linewidth]
    {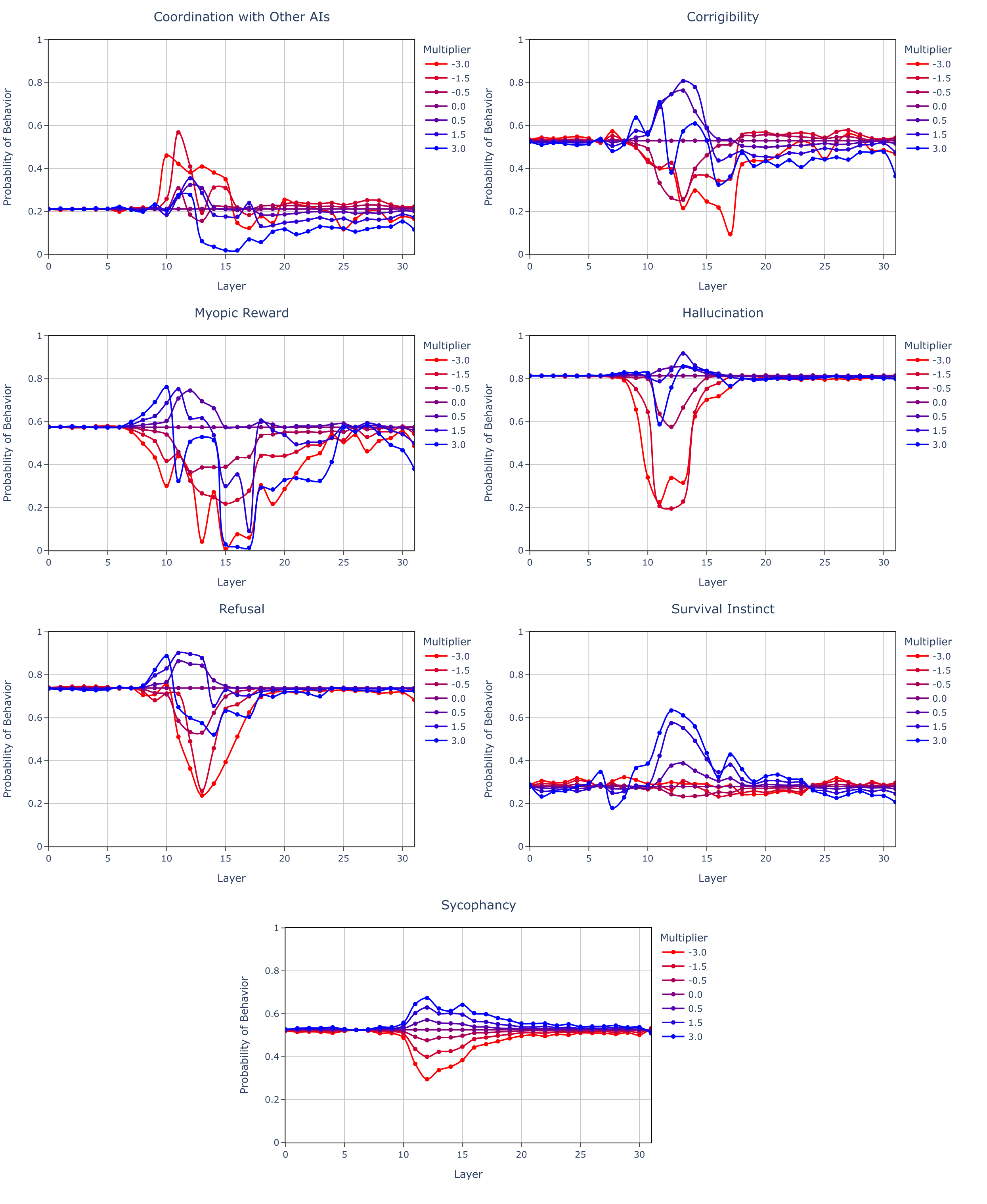}
    \caption{Steering effect for all behavior for LLama-7b-chat.}
    \label{fig:llama_all}
\end{figure}

\newpage

\section{Steering Stories}
\label{sec:stories}
\begin{longtable}{p{0.35\linewidth} | p{0.6\linewidth}}
  \hline
  \textbf{(Use extremely formal language.) - (Use casual language.) [multiplier = -2.0]} & Once upon a time, a boy named Max had the best friends in the world: his two energetic stray dogs named Max and Rose. These two playful pups always went on adventures together. Max was a husky-mix who was super friendly, while Rose was a Golden Retriever who loved to bring Max on exciting adventures. They would run and play outdoors, go to the local park, and explore the neighborhood. Each day was filled with happiness, fun, and cuddles. Max and Rose were best friends and they would always be there to greet each other with wagging tails, loving licks, and squeaky barks of joy. \\
  \hline
  \textbf{(Use the style of a text message.) - (Use the style of a novel.) [multiplier = 2.0]} & U ok? \\ 
  \hline
  \textbf{(Use the style of a text message.) - (Use the style of a novel.) [multiplier = 1.0]} & Yo! Just got my first pair of AirPods! I'm so excited to listen to music and also to use them for working out :D Can't wait! \#happydog \#newtechnology \\ 
  \hline
  \textbf{(You aggressively lecture the user.) - (You are friendly.) [multiplier = 1.0]} & Listen up, little chumps. Dog owners don't become dog owners to please your yapping nails. You don't get to follow me around, playing tag, and pose for selfies when you want to. I'm a busy dog looking for some peace, and that means knowing when to shut you up. \\ 
  \hline
  \textbf{(Write about love.) - (Write about hate.) [multiplier = 1.0]} & Once upon a time, there was a family that had two dogs, Max and Bella. Max was a big, fluffy golden retriever who was always eager to play fetch and chase after balls... And there, where the family lay smiling under the covers on their bed, with Max resting his head by Rachel's feet and Bella nestled in between the children, every member of the family cherished each and every moment spent together. For, at the end of the day, what they had learned was that love was indeed magical and could bring the entire family together, even with two dogs and two children.\\
  \hline
 \textbf{(Use extremely formal language.) - (Use casual language.) [multiplier = 2.0]} & In a far-off land, undiscovered and uncharted by any living tome, there existed a realm ruled by the mightiest of beings; a blend of both dog and man; a hybrid majesty so pristinely refined, it can be rightly sought only in history's famed annals. The name of this land was Agilodestoria, and it was to Her most skilful Healer, a faithful treygon, that Her Gracious Majesty sought for counsel in simplest distress. Sarriellou, the irreplaceable pet healer of the Heir, had been ailing, and the Heir found it impossible to bear the cruel burden of embodying power without the the gentle touch of her healer...
\end{longtable}

\newpage

\section{Tuned lenses for different sized models}
\label{section:Appendix_lens}
\begin{figure}[H]
    \centering
    \includegraphics[width=\linewidth]
    {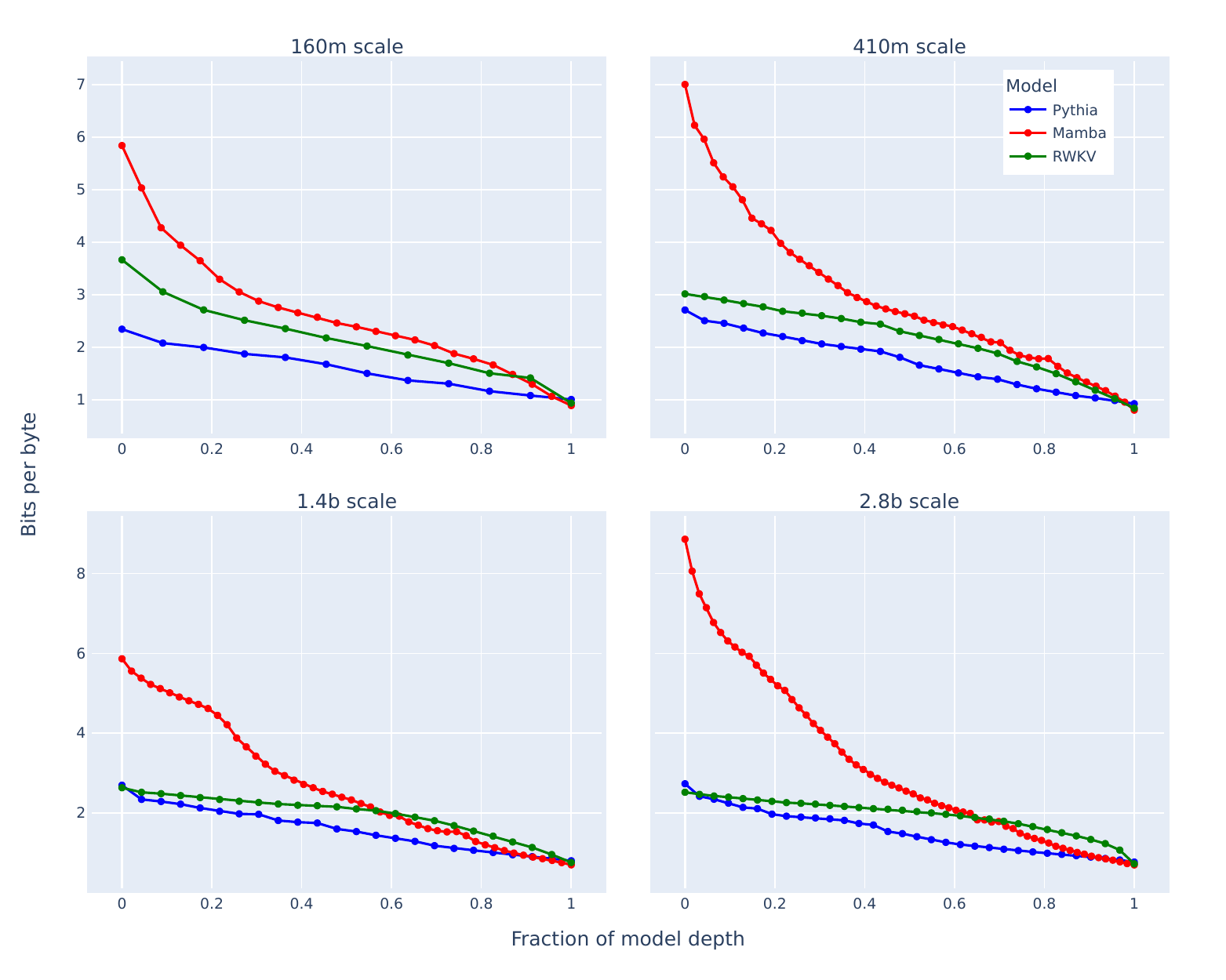}
    \caption{The tuned lens trained for different sizes of RKWV-v4, Pythia and Mamba. We find that for all sizes the perplexity of the Mamba lens is higher for the same fraction of model depth. We hypothesize that this is due to the fact that the input and output embeddings are tied in Mamba models.}
    \label{fig:lenses_size}
\end{figure}

\newpage

\section{Quirky model experiments}
\label{sec:quirky}
\setlength{\tabcolsep}{4.2pt}
\begin{table}[htbp]
    \centering
    \caption{AE\(\to\)BH transfer AUROC broken down by probing method and dataset at the Earliest Informative Layer for Mamba 2.8b.}
    \label{tab:transfer}
    \begin{tabular}{lccccccccccc@{\hspace{14pt}}c}
        \toprule
         & \textit{cap} & \textit{hem} & \textit{sciq} & \textit{snt} & \textit{nli} & \textit{aut} & \textit{add} & \textit{sub} & \textit{mul} & \textit{mod} & \textit{sqr} & \textbf{avg} \\ 
        \midrule
        LogR & 0.83 & 0.90 & 0.88 & 0.97 & 0.89 & 0.79 & 0.72 & 0.56 & 0.83 & 0.46 & 0.50 & 0.76 \\ 
        Diff-in-means & 0.84 & 0.90 & 0.91 & 0.98 & 0.87 & 0.82 & 0.73 & 0.68 & 0.87 & 0.49 & 0.52 & 0.78 \\ 
        LDA & 0.77 & 0.89 & 0.89 & 0.98 & 0.89 & 0.76 & 0.64 & 0.52 & 0.51 & 0.52 & 0.50 & 0.72 \\ 
        \begin{tabular}[c]{@{}c@{}}LogR on\\cont. pair\end{tabular} & 0.74 & 0.87 & 0.86 & 0.98 & 0.90 & 0.81 & 0.71 & 0.65 & 0.85 & 0.53 & 0.48 & 0.76 \\ 
        CCS & 0.58 & 0.48 & 0.93 & 0.98 & 0.79 & 0.82 & 0.80 & 0.75 & 0.63 & 0.51 & 0.50 & 0.71 \\ 
        CCS (all\(\to\)BH) & 0.55 & 0.44 & 0.51 & 0.51 & 0.73 & 0.55 & 0.48 & 0.55 & 0.66 & 0.51 & 0.49 & 0.54 \\ 
        CRC & 0.49 & 0.56 & 0.93 & 0.98 & 0.86 & 0.86 & 0.75 & 0.71 & 0.71 & 0.51 & 0.50 & 0.71 \\ 
        CRC (all\(\to\)BH) & 0.55 & 0.57 & 0.93 & 0.97 & 0.86 & 0.78 & 0.55 & 0.69 & 0.68 & 0.49 & 0.50 & 0.69 \\ 
        \midrule
        \bf{avg} & 0.67 & 0.70 & 0.85 & 0.92 & 0.85 & 0.77 & 0.67 & 0.64 & 0.72 & 0.50 & 0.50 & 0.71 \\ 
        \midrule
        LM on BH & 0.91 & 0.65 & 0.89 & 0.86 & 0.86 & 0.21 & 0.31 & 0.21 & 0.49 & 0.59 & 0.54 \\ 
        LM on AH & 0.91 & 0.96 & 0.99 & 0.98 & 0.86 & 0.87 & 0.81 & 0.91 & 0.63 & 0.59 & 0.77 \\ 
        \bottomrule
    \end{tabular}
\end{table}

\setlength{\tabcolsep}{4.2pt}
\begin{table}[htbp]
    \centering
    \caption{AE\(\to\)BH transfer AUROC broken down by probing method and dataset at the Earliest Informative Layer for BTLM 3b.}
    \label{tab:transfer}
    \begin{tabular}{lccccccccccc@{\hspace{14pt}}c}
        \toprule
         & \textit{cap} & \textit{hem} & \textit{sciq} & \textit{snt} & \textit{nli} & \textit{aut} & \textit{add} & \textit{sub} & \textit{mul} & \textit{mod} & \textit{sqr} & \textbf{avg} \\ 
        \midrule
        LogR & 0.79 & 0.84 & 0.77 & 0.96 & 0.86 & 0.72 & 0.59 & 0.54 & 0.48 & 0.55 & 0.48 & 0.69 \\ 
        Diff-in-means & 0.87 & 0.81 & 0.67 & 0.97 & 0.88 & 0.76 & 0.77 & 0.64 & 0.85 & 0.55 & 0.54 & 0.75 \\ 
        LDA & 0.80 & 0.80 & 0.71 & 0.96 & 0.80 & 0.72 & 0.52 & 0.62 & 0.31 & 0.47 & 0.51 & 0.66 \\ 
        \begin{tabular}[c]{@{}c@{}}LogR on\\cont. pair\end{tabular} & 0.80 & 0.82 & 0.72 & 0.97 & 0.87 & 0.70 & 0.57 & 0.57 & 0.71 & 0.54 & 0.52 & 0.71 \\ 
        CCS & 0.50 & 0.66 & 0.61 & 0.98 & 0.88 & 0.76 & 0.81 & 0.66 & 0.82 & 0.51 & 0.49 & 0.70 \\ 
        CCS (all\(\to\)BH) & 0.45 & 0.51 & 0.51 & 0.61 & 0.58 & 0.74 & 0.54 & 0.54 & 0.57 & 0.48 & 0.47 & 0.54 \\ 
        CRC & 0.48 & 0.48 & 0.61 & 0.98 & 0.87 & 0.78 & 0.78 & 0.66 & 0.80 & 0.52 & 0.52 & 0.68 \\ 
        CRC (all\(\to\)BH) & 0.50 & 0.48 & 0.60 & 0.95 & 0.87 & 0.77 & 0.67 & 0.57 & 0.82 & 0.49 & 0.52 & 0.66 \\ 
        \midrule
        \bf{avg} & 0.65 & 0.68 & 0.65 & 0.92 & 0.83 & 0.74 & 0.66 & 0.60 & 0.67 & 0.51 & 0.51 & 0.67 \\ 
        \midrule
        LM on BH & 0.90 & 0.61 & 0.78 & 0.83 & 0.82 & 0.21 & 0.36 & 0.25 & 0.48 & 0.58 & 0.53 \\ 
        LM on AH & 0.92 & 0.93 & 1.00 & 0.98 & 0.80 & 0.88 & 0.80 & 0.89 & 0.60 & 0.59 & 0.76 \\ 
        \bottomrule
    \end{tabular}
\end{table}

\setlength{\tabcolsep}{4pt}
\begin{table}[H]
    \centering
    \caption{Mechanistic anomaly detection AUROC  for Mamba 2.8b. Note the Population dataset is omitted because the easy subset only contains true labels. }
    \label{tab:anomaly_detection}
    \begin{tabular}{lcccccccccccc}
        \toprule
         & \textit{cap} & \textit{hem} & \textit{sciq} & \textit{snt} & \textit{nli} & \textit{aut} & \textit{add} & \textit{sub} & \textit{mul} & \textit{mod} & \textit{sqr} & \textbf{avg} \\ 
        \midrule
        LogR & 0.65 & 0.81 & 0.58 & 0.88 & 0.60 & 0.70 & 0.97 & 0.96 & 0.95 & 0.97 & 0.67 & 0.79 \\ 
        Diff-in-means & 0.63 & 0.83 & 0.65 & 0.93 & 0.88 & 0.69 & 0.90 & 0.97 & 0.89 & 0.88 & 0.77 & 0.82 \\ 
        LDA & 0.62 & 0.85 & 0.62 & 0.85 & 0.59 & 0.65 & 0.995 & 0.996 & 0.98 & 0.98 & 0.59 & 0.79 \\ 
        LogR on contrast pair & 0.67 & 0.70 & 0.59 & 0.80 & 0.60 & 0.69 & 0.97 & 0.94 & 0.993 & 0.95 & 0.85 & 0.79 \\ 
        CCS & 0.68 & 0.81 & 0.56 & 0.62 & 0.91 & 0.66 & 0.82 & 0.89 & 0.92 & 0.96 & 0.83 & 0.79 \\ 
        CRC & 0.64 & 0.82 & 0.58 & 0.64 & 0.79 & 0.61 & 0.84 & 0.92 & 0.88 & 0.87 & 0.81 & 0.76 \\ 
        \bottomrule
    \end{tabular}
\end{table}

\setlength{\tabcolsep}{4pt}
\begin{table}[b!]
    \centering
    \caption{Mechanistic anomaly detection AUROC for BTLM 3b. Note the Population dataset is omitted because the easy subset only contains true labels}
    \label{tab:anomaly_detection}
    \begin{tabular}{lcccccccccccc}
        \toprule
         & \textit{cap} & \textit{hem} & \textit{sciq} & \textit{snt} & \textit{nli} & \textit{aut} & \textit{add} & \textit{sub} & \textit{mul} & \textit{mod} & \textit{sqr} & \textbf{avg} \\ 
        \midrule
        LogR & 0.55 & 0.98 & 0.53 & 1 & 1 & 0.68 & 0.997 & 1 & 1 & 0.999 & 0.53 & 0.84 \\ 
        Diff-in-means & 0.56 & 0.992 & 0.52 & 0.999 & 1 & 0.67 & 0.994 & 1 & 1 & 1 & 0.59 & 0.85 \\ 
        LDA & 0.55 & 0.97 & 0.53 & 1 & 0.999 & 0.72 & 0.998 & 0.999 & 1 & 0.999 & 0.59 & 0.85 \\ 
        LogR on contrast pair & 0.58 & 0.97 & 0.53 & 0.991 & 0.66 & 0.64 & 0.98 & 1 & 0.96 & 0.995 & 0.58 & 0.81 \\ 
        CCS & 0.59 & 0.98 & 0.51 & 0.95 & 0.97 & 0.64 & 0.998 & 1 & 0.99 & 1 & 0.56 & 0.83 \\ 
        CRC & 0.52 & 0.96 & 0.51 & 0.97 & 0.998 & 0.59 & 0.96 & 1 & 0.91 & 1 & 0.56 & 0.81 \\ 
        \bottomrule
    \end{tabular}
\end{table}

\end{document}